\documentclass[11pt]{article}

\usepackage[preprint]{acl}
\usepackage{times}
\usepackage{latexsym}
\usepackage[T1]{fontenc}
\usepackage[utf8]{inputenc}
\usepackage{microtype}
\usepackage{booktabs}
\usepackage{graphicx}
\usepackage{float}

\usepackage{amsmath,amsfonts,bm}

\def\eqref#1{equation~\ref{#1}}

\def\1{\bm{1}}

\DeclareMathAlphabet{\mathsfit}{\encodingdefault}{\sfdefault}{m}{sl}
\SetMathAlphabet{\mathsfit}{bold}{\encodingdefault}{\sfdefault}{bx}{n}

\hypersetup{
  pdftitle={ExBind: A Controlled Diagnostic Benchmark for Visual-to-Executable Correspondence},
  pdfauthor={Ziqian Wang; Yuxiao Cheng; Tingxiong Xiao; Jinli Suo},
  pdfsubject={Benchmark and diagnostic evaluation of executable multimodal reference binding},
  pdfkeywords={multimodal models, reference binding, visual grounding, benchmark, diagnostic evaluation}
}

\title{ExBind: A Controlled Diagnostic Benchmark for Visual-to-Executable Correspondence}

\author{Ziqian Wang \quad Yuxiao Cheng \quad Tingxiong Xiao \quad Jinli Suo\textsuperscript{*}\\
Department of Automation, Tsinghua University\\
Beijing 100084, China\\
\textsuperscript{*}Corresponding author: \texttt{jlsuo@tsinghua.edu.cn}}
\begin{document}

\maketitle

\begin{abstract}
Iterative visual-feedback coding and editing systems repeatedly translate between rendered artifacts and executable program structure: after a model or user identifies a visible target, the system must resolve that referent to the precise object that can be edited. A failure at this correspondence layer can send an otherwise sensible patch to the wrong DOM node, SVG child, graph endpoint, hierarchy member, or table cell, but end-to-end edit success does not reveal whether the cause was localization, correspondence, code generation, execution, or accumulated history. ExBind isolates visual/semantic-to-executable correspondence as a controlled diagnostic target between semantic localization and action execution. It samples representation-independent latent binding instances and compiles them into SVG, DOM, canvas, tree, graph, and table cases with invertible mappings to surface executable references. Models emit only the final strict reference; the evaluator maps it back to latent structure and reports deterministic structural constraints, without requiring model-generated reasoning traces. The primary release contains a 250-case broad suite, a disjoint 240-case targeted suite, and 50 paired latent groups. On the broad suite, Qwen2.5-VL-3B attains 98.4\% candidate validity but 76.4\% exact accuracy, whereas Qwen3-VL-4B attains 100.0\% validity and 98.8\% exact accuracy. In the targeted table diagnostic, every Qwen2.5-VL-3B residual error is a valid correct-row/wrong-column selection under the declared decomposition. Candidate-order perturbations change case-level outcomes while preserving this localized error form. ExBind is a controlled diagnostic instrument for executable correspondence, rather than a population-scale model leaderboard or an end-to-end editing benchmark. Code and benchmark records are publicly available at \url{https://github.com/Daerwang2020/Exbind} and \url{https://huggingface.co/datasets/Ziqianwwww/ExBind}, respectively.
\end{abstract}

\section{Introduction}

Iterative visual-feedback coding and editing systems repeatedly move between a rendered artifact and the program that controls it. A typical loop is:
\begin{center}
\small
code $\rightarrow$ render $\rightarrow$ inspect visual feedback $\rightarrow$ identify a referent $\rightarrow$ resolve its executable object $\rightarrow$ patch $\rightarrow$ render again.
\end{center}
Suppose a user asks the system to ``move the right output box slightly down.'' The visible referent may be clear, but executing the request requires identifying which DOM node, SVG child, component, or code-controlled object denotes that box. If the correspondence is wrong, a sensible patch can be applied to the wrong object; repeated revisions may then compound the discrepancy.

This motivates the central question of ExBind:
\begin{center}
\large\boxed{\text{Where does executable binding first fail?}}
\end{center}
Here ``first'' means the first violated structural constraint under a declared evaluator decomposition, not the first step in a model's hidden reasoning. ExBind isolates the middle layer in the following decomposition:
\begin{center}
\small
semantic/visual localization $\rightarrow$ \textbf{structured executable correspondence} $\rightarrow$ action or code execution.
\end{center}
An end-to-end editing outcome conflates these layers. A failed patch may reflect incorrect localization, an incorrect referent-to-program correspondence, faulty patch generation, execution failure, or accumulated errors from earlier turns. ExBind therefore does not benchmark the full multi-turn loop; it makes the correspondence layer measurable in isolation.

A final executable failure is an observation, not a diagnosis. Consider the instruction ``select the status cell for the project owned by Team B.'' The correct cell requires resolving the project row, the requested attribute column, and their intersection. Selecting the owner cell from the correct row is a valid executable output, yet it differs fundamentally from returning an unknown identifier or a cell from the wrong project. Analogous distinctions arise when a model identifies the correct visual object but the wrong executable child, chooses the correct graph nodes but reverses source and target roles, or selects a valid DOM element from the wrong hierarchy. Thus candidate validity, binding correctness, and failure diagnosis are distinct quantities:
\begin{center}
\small
\(\text{validity} \neq \text{binding correctness} \neq \text{failure diagnosis}.\)
\end{center}

Existing grounding evaluations usually stop at a region, coordinate, element, or action \citep{mdetr,glip,groundingdino,seeclick,seeact,ferretui,omniparser}. They measure semantic localization, an essential prerequisite, but do not generally make the subsequent referent-to-executable structural correspondence the primary diagnostic object. Diagnostic visual reasoning benchmarks similarly show the value of structured annotations for separating conflated errors \citep{clevr,gqa}. In executable environments, the relevant constraints are often directly available: entities have attributes and parents, graph endpoints have ordered roles, DOM elements lie in a hierarchy, and table cells are row--column intersections.

We introduce ExBind, a controlled diagnostic instrument for structured executable correspondence. Each instance begins with a representation-independent latent binding object containing entities, relations, hierarchy, attributes, roles, and a canonical target. Format-specific compilers materialize the same binding object as SVG, DOM, canvas, tree, graph, or table observations with surface-specific executable references. Visual correspondence cases instantiate the motivating rendered setting more directly; controlled structural cases use serialized or symbolic surfaces to isolate hierarchy, role, row/column, and executable-identity constraints. The model still returns only the reference required by the environment. The evaluator deterministically maps the prediction back into the latent binding space, producing a full set of violated applicable constraints and, where the family has a declared structural order, a concise first-violation summary. This construction supports comparison across surfaces without requiring their surface identifiers to match.

The primary release contains a 250-case broad suite, a disjoint 240-case targeted structural suite, and 50 paired latent groups. Its controlled size follows from the need for known latent targets, unique executable gold, deterministic mappings, structural-oracle recovery, and shortcut checks. The goal is diagnostic identifiability and measurement reliability, not population-scale ranking. The inherited 5.9K SVG/DOM inventory is retained as provenance and a cached-output bridge, not pooled into the primary evaluation.

Frozen-model evaluation illustrates the diagnostic use case. On the broad suite, Qwen2.5-VL-3B returns a valid candidate on 98.4\% of cases but is exact on 76.4\%; Qwen3-VL-4B reaches 98.8\% exact accuracy, leaving only three valid-but-wrong residual cases under the same vocabulary. In the targeted table suite, Qwen2.5-VL-3B is valid on all 120 cases yet exact on 63, and every residual error is localized to the column-selection constraint after row resolution. Candidate-order perturbations change error frequency and case identity, but all residual errors preserve this correct-row/wrong-column form. Thus outcome stability and diagnostic-type stability are distinct measurement properties. The hierarchy-depth control remains at ceiling for both models, delimiting rather than expanding the benchmark claim.

We make three contributions:
\begin{enumerate}
    \item \textbf{Executable correspondence as an isolated diagnostic target.} We identify and formalize the referent-to-executable correspondence layer between semantic/visual localization and action or code execution in visual-feedback editing workflows.
    \item \textbf{Controlled latent-to-surface construction.} We introduce ExBind, which compiles representation-independent latent binding instances into heterogeneous executable surfaces while retaining deterministic mappings needed to diagnose valid-but-wrong predictions.
    \item \textbf{Residual structure and interface sensitivity.} We show that candidate validity can mask executable mismatches, that residual profiles differ across frozen checkpoints, and that interface serialization can change case-level outcomes while preserving the localized structural failure form.
\end{enumerate}

\begin{figure*}[t]
\centering
\includegraphics[width=0.98\linewidth]{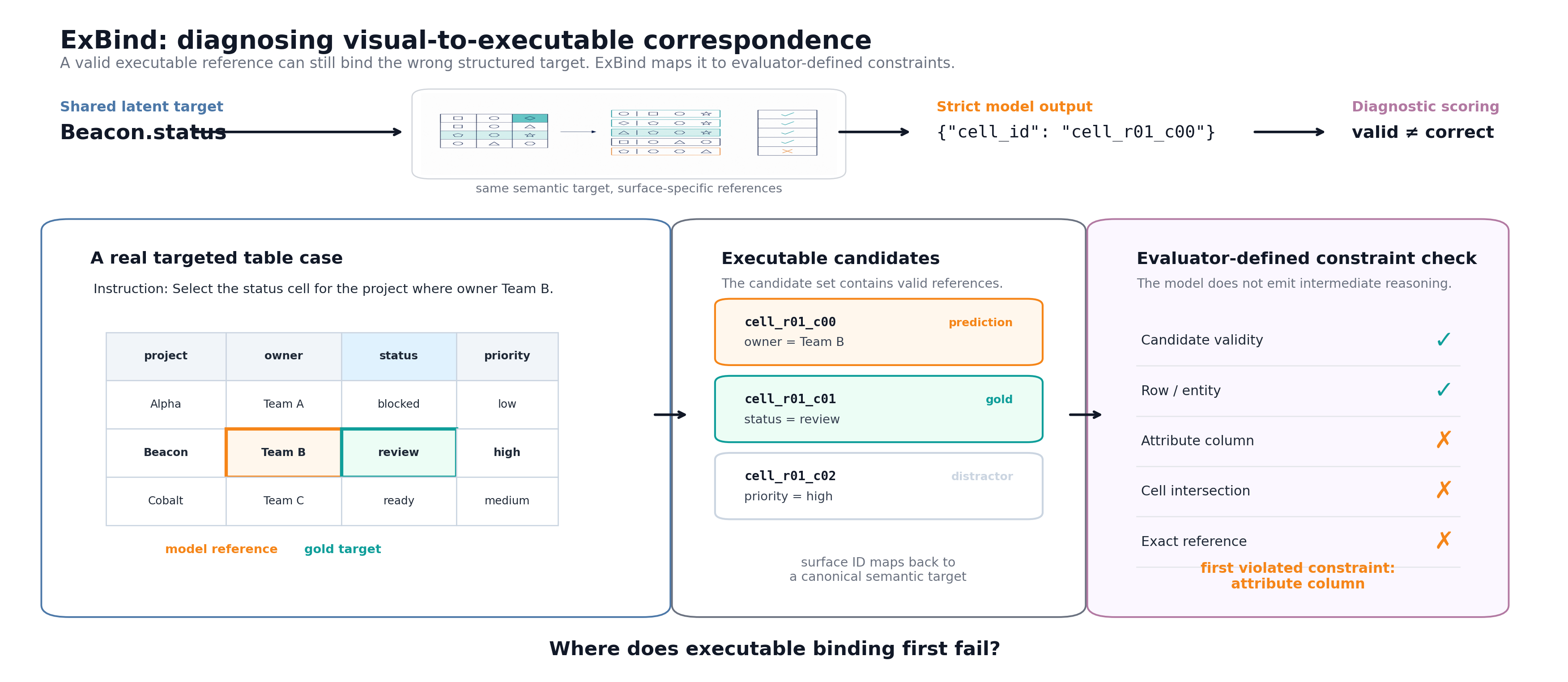}
\caption{ExBind isolates visual-to-executable correspondence from a larger visual-feedback editing loop. A shared latent target is compiled into surface-specific observations and candidate inventories; the model emits only a strict reference; and the evaluator maps that reference back to the latent target and checks deterministic constraints. The running example is an actual targeted table case: \texttt{cell\_r01\_c00} is a valid candidate in the correct row, yet it selects the owner rather than the status column. The first-violation summary is evaluator-defined rather than a model reasoning trace.}
\label{fig:overview}
\end{figure*}

\section{Related Work}

ExBind is positioned between three layers that are often evaluated together but are scientifically distinct:
\begin{center}
\small
semantic localization \(\rightarrow\) structured executable binding \(\rightarrow\) action execution.
\end{center}
The benchmark evaluates the middle layer: after a semantic or visual referent is identified, which environment-specific executable reference denotes it? Its diagnostic labels are mechanically recoverable from executable environment structure and the latent target; they are neither model-generated reasoning traces nor a process-supervision signal.

\paragraph{Referring and visual grounding.}
Referring-expression comprehension and phrase grounding align natural language with image regions or objects. MDETR jointly models text and images for text-conditioned object detection and referring-expression tasks, while GLIP and Grounding DINO extend grounding toward large-scale and open-vocabulary recognition \citep{mdetr,glip,groundingdino}. Multimodal LLMs such as Ferret further support referring and grounding over points, boxes, and free-form regions \citep{ferret}. These methods address a prerequisite of ExBind---identifying the intended visual referent---but their typical target remains a region or object representation. ExBind studies the subsequent structured resolution from an already specified or recoverable referent to an environment-specific executable reference, where hierarchy, ordered roles, or symbolic identity may matter even after the relevant visual entity has been recognized. The distinction is therefore not that prior grounding work lacks structure, but that referent-to-executable correspondence is not generally its primary diagnostic object.

\paragraph{GUI grounding and computer-use agents.}
Grounding has become a central component of multimodal computer-use agents. SeeClick introduces ScreenSpot and shows that improved GUI grounding is associated with improved downstream agent performance, while SeeAct demonstrates a large gap between strong multimodal planning and automatic grounding on web interfaces \citep{seeclick,seeact}. Ferret-UI, ScreenAI, OmniParser, and subsequent GUI-specific models improve screen understanding, interactable-element parsing, or action grounding \citep{ferretui,screenai,omniparser}. More recent benchmarks increase the realism and diversity of grounding evaluation: ScreenSpot-Pro targets high-resolution professional software, WinSpot focuses on Windows applications, and UI-Vision evaluates element grounding, layout grounding, and action prediction in desktop environments \citep{screenspotpro,winspot,uivision}. MMBench-GUI further organizes GUI-agent evaluation hierarchically from content understanding and element grounding to task automation and collaboration \citep{mmbenchgui}. ExBind is complementary to this progression: rather than extending the environment or agent horizon, it isolates the binding decision itself and decomposes it into verifiable structural constraints.

\paragraph{Diagnostic and compositional evaluation.}
ExBind also follows the tradition of diagnostic benchmarks designed to separate error sources that aggregate task metrics conflate. CLEVR introduces controlled visual scenes and detailed annotations to diagnose compositional visual reasoning, explicitly motivated by the difficulty of identifying which capability causes a VQA failure \citep{clevr}. GQA moves this idea to real images using scene graphs and functional programs, and supplements answer accuracy with consistency, grounding, and plausibility analyses \citep{gqa}. These works diagnose visual reasoning requirements underlying question answering. ExBind applies the same general principle to executable reference resolution, but its diagnostic annotations are derived from executable binding structure rather than from functional programs or model-generated reasoning. This distinction separates ExBind from process supervision and chain-of-thought verification: the evaluated model emits only its final executable reference, and all intermediate labels are deterministic benchmark annotations \citep{processsupervision,cot}.

\paragraph{Structured visual and semi-structured artifacts.}
Web pages, user interfaces, diagrams, and tables couple visual layout with an underlying symbolic structure. WebSRC evaluates structural reading comprehension using webpages with screenshots, HTML, and metadata, while Pix2Struct exploits the alignment between webpage screenshots and HTML as a general pretraining signal for visually situated language \citep{websrc,pix2struct}. ScreenAI similarly studies UI and infographic understanding through structured screen annotation, and ChartQA evaluates visual and logical reasoning over charts and their underlying data \citep{screenai,chartqa}. ExBind uses this dual surface/structure property for a different purpose: a shared latent binding vocabulary is instantiated across heterogeneous representations and used to score where a prediction ceases to match the executable structure.

Table~\ref{tab:benchmark_positioning} summarizes the distinction at the benchmark level. A positive entry means that the benchmark makes the property an explicit evaluation object, rather than merely allowing a model to consume structured input.

\begin{table*}[t]
\caption{Positioning against representative grounding, diagnostic reasoning, GUI, and structured-artifact benchmarks. ExBind's distinguishing combination is a surface-specific executable target, evaluator-defined structural labels, and paired latent cases across formats.}
\label{tab:benchmark_positioning}
\centering
\scriptsize
\begin{tabular*}{\linewidth}{@{\extracolsep{\fill}}p{0.21\linewidth}ccccc@{}}
\toprule
Benchmark family & Executable target & Structured latent/gold & Multiple formats & Paired cross-format cases & Stage-localized errors\\
\midrule
RefCOCO / phrase grounding~\citep{refcoco,mdetr} & no & partial & no & no & no\\
CLEVR / GQA~\citep{clevr,gqa} & no & yes & no & no & yes\\
GUI grounding~\citep{seeclick,screenspotpro,uivision} & partial & partial & partial & no & no\\
WebSRC / Pix2Struct~\citep{websrc,pix2struct} & no & yes & screenshot+source & no & no\\
ChartQA~\citep{chartqa} & no & partial & chart formats & no & no\\
\textbf{ExBind-v2} & \textbf{yes} & \textbf{yes} & \textbf{yes} & \textbf{yes} & \textbf{yes}\\
\bottomrule
\end{tabular*}
\end{table*}

\section{Problem Formulation}

We represent one latent binding instance as
\begin{equation}
G=(V,E,A,H,R,Y),
\end{equation}
where $V$ is a set of entities, $E$ contains structural or graph edges, $A$ stores attributes, $H$ stores parent-child structure, $R$ stores semantic or spatial relations, and $Y$ is the target binding specification. For the table example, $V$ contains project rows, $A$ contains owner/status/priority values, and $Y$ specifies the status attribute of the row whose owner is Team B.

A representation compiler $K_f$ materializes this latent object into a format-specific case:
\begin{equation}
X_f=K_f(G)=(O_f,I,C_f,Y_f,M_f).
\end{equation}
Here $O_f$ is the observation, $I$ is the instruction, $C_f$ is the executable candidate set, $Y_f$ is the surface gold reference, and $M_f$ maps latent objects to surface references. The table compiler may map the same latent target to \texttt{cell\_r01\_c01}; a graph compiler may instead expose an ordered endpoint pair; a DOM compiler may expose a \texttt{dom\_id}.

The model prediction $\hat{y}_f$ is a strict structured reference, e.g. \texttt{\{target\_id: ...\}}, \texttt{\{source\_id: ..., target\_id: ...\}}, \texttt{\{dom\_id: ...\}}, or \texttt{\{cell\_id: ...\}}. ExBind maps $\hat{y}_f$ through $M_f^{-1}$ when possible and computes a deterministic diagnostic score vector $s(\hat{y}_f,G)$ together with the full violated-constraint set
\begin{equation}
V(\hat{y}_f,G)=\{d \in D_f : s_d(\hat{y}_f,G)=0\},
\end{equation}
where $D_f$ is the set of applicable structural constraints for format $f$. The model is never asked for chain-of-thought; intermediate labels are benchmark annotations derived from the latent record and candidate inventory. For task families with a declared structural order, we additionally report the first member of $V$ under that order as a compact diagnostic summary. This ordering is an evaluator-defined decomposition of the executable target. It does not imply that the model internally performs these operations sequentially, and it is not a causal account of model computation.

For the running table example, the ordered stages are:
\begin{center}
\small
row/entity $\rightarrow$ attribute column $\rightarrow$ cell intersection $\rightarrow$ executable reference.
\end{center}
Other families use different applicable stages. A graph endpoint task uses identity, relation, ordered role, and exact endpoint-pair reference. A tree/DOM task uses identity, hierarchy, role, and exact node reference. Not-applicable dimensions are reported with null denominators rather than counted as failures.

\paragraph{Benchmark unit and evaluator boundary.}
The unit of evaluation is one compiled case $x_f=(O_f,I,C_f,Y_f,M_f)$ together with its latent record $G$. The model receives the observation, instruction, and candidate inventory and returns one surface executable reference; it is not asked to expose intermediate reasoning. The evaluator then applies the fixed mapping $M_f^{-1}$ and checks the applicable structural constraints. Thus a stage-localized structural mismatch is a property of the pair $(\hat{y}_f,G)$ under the benchmark decomposition, not a claim about the order or content of hidden model computations. Cases are scored independently, and cross-format consistency is computed only within explicitly paired latent groups.

\section{Benchmark Construction and Validation}

\subsection{Benchmark Composition}

ExBind-v2 separates the primary evaluation protocol from inherited diagnostic resources. The primary protocol consists of a 250-case broad diagnostic suite and a 240-case targeted structural diagnostic. These suites are deliberately controlled samples: their purpose is to make correspondence constraints identifiable and to support reliable stage-level comparison under fixed interfaces. The inherited 5.9K ExBind-Bench SVG/DOM inventory is preserved as a legacy resource for provenance, historical characterization, and cached-output bridge analysis; it is not pooled into the primary evaluation split. Table~\ref{tab:components} summarizes this paper-facing composition.

\begin{table*}[t]
\caption{Paper-facing benchmark components. Primary claims are based on the new ExBind-v2 core evaluation; legacy resources support provenance and bridge analysis.}
\label{tab:components}
\centering
\small
\begin{tabular*}{\linewidth}{@{\extracolsep{\fill}}lrrll@{}}
\toprule
Component & N & New? & Model inference? & Primary claim?\\
\midrule
Broad diagnostic suite & 250 & yes & yes & yes\\
Targeted table diagnostic & 120 & yes & yes & yes\\
Targeted hierarchy diagnostic & 120 & yes & yes & supporting\\
Cross-format paired groups & 50 & yes & yes & supporting\\
LLaVA-Phi-3-mini supporting pilot & 250+120 & no & yes & supporting\\
Legacy bridge subset & 300 & no & cached & supporting\\
Inherited ExBind-Bench inventory & 5,900 & no & historical & no\\
\bottomrule
\end{tabular*}
\end{table*}

\begin{table}[t]
\caption{Supporting non-Qwen evaluation under the same ordinary strict-reference protocol. These results are reported as a cross-family diagnostic profile, not pooled into the primary Qwen comparison.}
\label{tab:nonqwen_support}
\centering
\small
\begin{tabular*}{\linewidth}{@{\extracolsep{\fill}}lrrr@{}}
\toprule
Split & $N$ & Valid & Exact\\
\midrule
Broad diagnostic & 250 & 220/250 (88.0) & 154/250 (61.6)\\
Targeted table & 120 & 67/120 (55.8) & 46/120 (38.3)\\
\bottomrule
\end{tabular*}
\end{table}

The broad suite instantiates canvas, graph, table, tree, SVG, and HTML-DOM cases. The targeted structural diagnostic is a separate 240-case split rather than an additional model-independent annotation of the broad split. Table~\ref{tab:split_composition} gives the paper-facing composition by representation. The table cases are the main stage-localization case study because they expose valid executable outputs that preserve the target row while selecting the wrong attribute column. The legacy bridge maps eligible cached EVG/SEB outputs into the same binding taxonomy without rerunning inference.

\begin{table*}[t]
\caption{Composition of the primary ExBind-v2 release. Broad and targeted cases are disjoint evaluation splits. Cross-format groups contain 25 SVG--canvas pairs and 25 tree--DOM pairs; they are used for consistency analysis rather than pooled exact accuracy.}
\label{tab:split_composition}
\centering
\small
\begin{tabular*}{\linewidth}{@{\extracolsep{\fill}}lrrrrrrr@{}}
\toprule
Split & SVG & Canvas & Tree & DOM & Graph & Table & Total\\
\midrule
Broad diagnostic & 25 & 50 & 50 & 25 & 50 & 50 & 250\\
Targeted structural & -- & -- & 120 & -- & -- & 120 & 240\\
Cross-format groups & 25 & 25 & 25 & 25 & -- & -- & 50 groups\\
\bottomrule
\end{tabular*}
\end{table*}

\subsection{Latent-to-Surface Construction}

Each generated case starts from a latent binding instance rather than from a surface format. Relational latents are compiled into canvas and SVG observations; hierarchy latents are compiled into tree and HTML-DOM observations; graph latents expose ordered endpoint candidates; and table latents expose row--column cell candidates. Rendered SVG and canvas cases instantiate visual correspondence more directly, whereas serialized DOM, tree, graph, and table cases hold structural variables explicit so that individual correspondence constraints can be identified. The compiler produces the observation, instruction, candidate inventory, surface gold, and latent-to-surface mapping. Appendix~\ref{app:reproducibility} gives the complete generation algorithm, record schema, prompt contract, and release commands.

\subsection{Representation Coverage}

Table~\ref{tab:formats} is paper-facing rather than an implementation status matrix. A checkmark means that the representation instantiates that dimension in evaluated cases; N/A means the dimension is not semantically part of that task family. Figure~\ref{fig:overview} shows the evaluation flow from latent instance to surface prediction and diagnostic scoring.

\begin{table*}[t]
\caption{Representation families and their verified binding coverage in the primary release. A checkmark denotes an applicable scored dimension; N/A denotes that the dimension is not part of that family.}
\label{tab:formats}
\centering
\small
\begin{tabular*}{\linewidth}{@{\extracolsep{\fill}}lcccccc@{}}
\toprule
Repr. & Identity & Attribute & Relation & Hierarchy & Ordered role & Structured selection\\
\midrule
SVG & yes & yes & yes & N/A & N/A & N/A\\
DOM & yes & N/A & N/A & yes & partial & N/A\\
Canvas2D & yes & yes & yes & N/A & N/A & N/A\\
Tree & yes & N/A & N/A & yes & yes & N/A\\
Graph & yes & N/A & yes & partial & yes & N/A\\
Table & yes & yes & N/A & partial & N/A & yes\\
\bottomrule
\end{tabular*}
\end{table*}

\subsection{Construct Validity}

Structured stage labels are useful only if they are mechanically grounded. ExBind therefore applies deterministic gates before model evaluation. The broad diagnostic suite and the targeted structural diagnostics require unique gold targets, non-empty candidate sets, deterministic latent-to-surface mappings, and a structural oracle that recovers the gold target from benchmark metadata. Table~\ref{tab:gates} summarizes the targeted diagnostic gates.

\begin{table*}[t]
\caption{Integrity gates for the 240-case targeted structural diagnostic.}
\label{tab:gates}
\centering
\begin{tabular*}{\linewidth}{@{\extracolsep{\fill}}lcc@{}}
\toprule
Gate & Result & Interpretation\\
\midrule
Unique gold target & 240/240 & no ambiguous gold cases\\
Structural oracle & 1.000 & latent metadata determines the gold\\
Target-ID leakage & 0.000 & instruction/candidates do not reveal target ID\\
Lexical-only baseline & 0.000 & token overlap alone does not solve cases\\
Type-only baseline & 0.000 & candidate type alone does not solve cases\\
Random-candidate expectation & 0.167 & mean chance accuracy over the 240 cases\\
Gold-rank variation & passed & candidate order is not a fixed gold-position proxy\\
Accepted records & 240/240 & no rejected diagnostic records\\
\bottomrule
\end{tabular*}
\end{table*}

\subsection{Shortcut Validity}

Shortcut checks are part of the benchmark acceptance criteria rather than post-hoc analysis. Target identifiers must not appear in instructions, candidate order must vary across accepted cases, lexical-only and type-only baselines must fail on the targeted diagnostic, and the structural oracle must remain exact. These controls do not prove that every shortcut is impossible; they define the minimum validity boundary under which structural diagnosis is interpretable.

\subsection{Measurement Reliability and Cross-Representation Validity}

Candidate serialization is an explicit protocol variable. For primary results, released records fix one candidate order; the matched permutation control treats that order as a nuisance variable of the executable interface because it changes access to an otherwise identical candidate set without changing the latent target. We therefore separate case-level outcome stability from diagnostic-type stability, and report candidate-order sensitivity as a measurement result rather than as a binding capability or causal explanation.

Cross-format groups provide a complementary validity check. Their shared latent target and surface mappings permit both semantic-consistency scoring and paired diagnostic transitions. A paired comparison can therefore distinguish a structural mismatch that persists across surfaces from one that appears only after surface compilation. The analysis is deliberately restricted to explicitly compatible compiler pairs; it does not assume that every latent instance can be rendered faithfully in every format.

Finally, we assess whether the controlled taxonomy can describe a small fixed sample of existing executable-binding errors. We manually adjudicate 20 cached errors drawn from EVG (8), SVGEditBench-derived resources (10), and MiniWoB DOM cases (2) against the stored instruction, candidate/gold record, and prediction. All 20 are covered by the existing taxonomy: 16 are wrong-identity errors, two are wrong-role/order errors, one is a wrong-relation error, and one is a wrong-type error. Five multi-edit source instructions are ambiguous at the task-unit level, but none requires a new taxonomy category. This is a sampled coverage check rather than an estimate of error prevalence in real agents; Appendix~\ref{app:existing_coverage} gives the annotation protocol and source breakdown.

\section{Evaluation Protocol}

We evaluate two primary frozen VLMs with the same strict reference-output contract: Qwen2.5-VL-3B and Qwen3-VL-4B. We additionally run LLaVA-Phi-3-mini as a supporting cross-family evaluation under the identical ordinary contract; it is not pooled into the primary Qwen comparison. The broad diagnostic suite contains 250 records and 50 cross-format latent groups. The targeted structural diagnostic contains 240 records: 120 hierarchy cases and 120 table cases. These are separate controlled diagnostic splits and are not pooled into a single leaderboard score. Cross-format groups are scored by canonical semantic consistency, while the targeted split is used for controlled failure diagnosis. The table diagnostic is reported by structural condition rather than as a calibrated scalar axis: compact single-condition tables, conjunctive-condition tables, and high-candidate duplicate-value tables. Candidate order is a fixed part of the released primary interface and a nuisance variable only in the matched permutation control.

Metrics are exact executable-reference accuracy, candidate validity, dimension accuracy, the full violated-constraint set, task-specific survival, first violated diagnostic constraint under the declared family order, and cross-format semantic consistency. Candidate validity asks whether every returned identifier belongs to the candidate inventory. Exact accuracy asks whether the executable surface reference exactly matches gold. Survival $S(k)$ is computed only for meaningful evaluator-defined stage orders within a task family. All proportions retain their case denominator; dimensions that do not apply are excluded rather than treated as incorrect.

We organize the empirical section around five scientific questions:
\begin{enumerate}
    \item Does validity hide executable-reference binding errors?
    \item Which diagnostic constraint is first violated when exact binding fails?
    \item Do residual failures differ across frozen VLMs?
    \item What structural mismatch pattern is exposed by the targeted table diagnostic?
    \item How reliable are diagnostic profiles under interface and cross-format controls?
\end{enumerate}

\section{Results}

\subsection{RQ1: Validity Is Not Binding Correctness}

Table~\ref{tab:main} shows that output validity and binding correctness measure different phenomena. Qwen2.5-VL-3B returns valid candidate references for 246/250 broad-suite cases (98.4\%) but is exact for only 191/250 (76.4\%). Most failures are therefore not malformed outputs. Qwen3-VL-4B reaches 250/250 validity and 247/250 exact accuracy; the remaining three errors are still valid candidate selections. The distinction is most informative when output formatting is mostly solved but binding remains imperfect.

\begin{table*}[t]
\caption{Broad diagnostic suite: validity and exact binding accuracy. Wilson 95\% intervals are shown for exact accuracy.}
\label{tab:main}
\centering
\begin{tabular*}{\linewidth}{@{\extracolsep{\fill}}lrrrrr@{}}
\toprule
Model & Cases & Valid & Exact & Exact 95\% CI & Errors after validity\\
\midrule
Qwen2.5-VL-3B & 250 & 0.984 & 0.764 & [0.708, 0.812] & 55\\
Qwen3-VL-4B & 250 & 1.000 & 0.988 & [0.965, 0.996] & 3\\
\bottomrule
\end{tabular*}
\end{table*}

\begin{table*}[t]
\caption{Primary Qwen results by representation. Values are valid/exact counts with percentages in parentheses. The first six rows partition the 250-case broad suite; the final two rows are the disjoint targeted structural split.}
\label{tab:format_results}
\centering
\scriptsize
\begin{tabular*}{\linewidth}{@{\extracolsep{\fill}}lrrrrr@{}}
\toprule
Representation / split & $N$ & \multicolumn{2}{c}{Qwen2.5-VL-3B} & \multicolumn{2}{c}{Qwen3-VL-4B}\\
 & & Valid & Exact & Valid & Exact\\
\midrule
Canvas2D & 50 & 50/50 (100.0) & 50/50 (100.0) & 50/50 (100.0) & 49/50 (98.0)\\
SVG & 25 & 25/25 (100.0) & 25/25 (100.0) & 25/25 (100.0) & 25/25 (100.0)\\
Tree & 50 & 50/50 (100.0) & 35/50 (70.0) & 50/50 (100.0) & 50/50 (100.0)\\
HTML-DOM & 25 & 25/25 (100.0) & 18/25 (72.0) & 25/25 (100.0) & 23/25 (92.0)\\
Graph & 50 & 46/50 (92.0) & 40/50 (80.0) & 50/50 (100.0) & 50/50 (100.0)\\
Table & 50 & 50/50 (100.0) & 23/50 (46.0) & 50/50 (100.0) & 50/50 (100.0)\\
\midrule
Targeted table & 120 & 120/120 (100.0) & 63/120 (52.5) & 120/120 (100.0) & 120/120 (100.0)\\
Targeted hierarchy & 120 & 120/120 (100.0) & 120/120 (100.0) & 120/120 (100.0) & 120/120 (100.0)\\
\bottomrule
\end{tabular*}
\end{table*}

\subsection{RQ2: ExBind Localizes Evaluator-Defined Structural Mismatches}

\begin{figure}[H]
\centering
\includegraphics[width=0.92\linewidth]{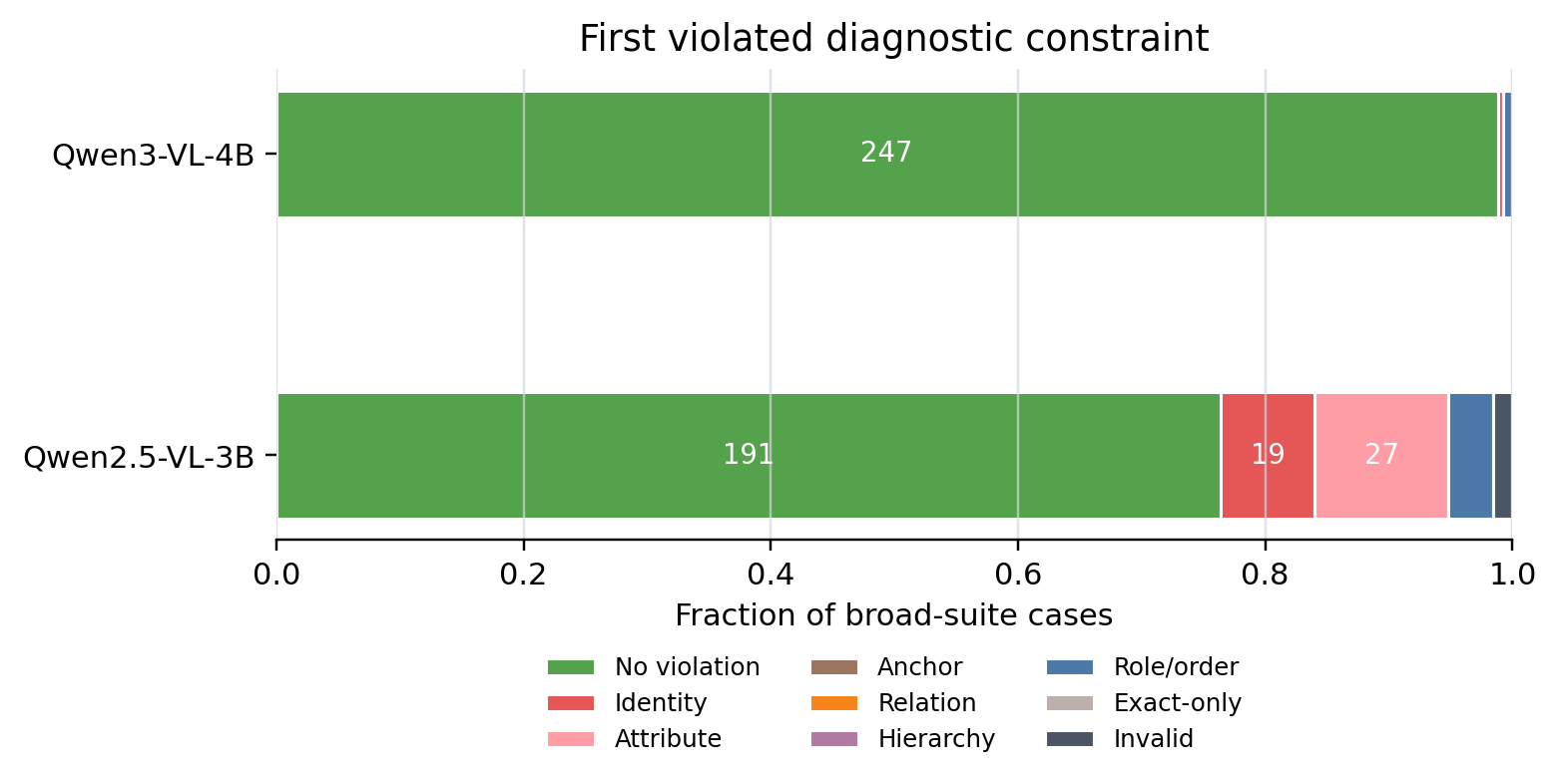}
\caption{Declared-order first violated diagnostic constraint on the broad diagnostic suite. Counts are computed by mapping each final executable prediction back to the latent target and checking evaluator-defined structural constraints; the released artifacts also retain the full violated-constraint set.}
\label{fig:constraint_profile}
\end{figure}

Dimension-level scores expose the structure hidden behind exact accuracy. Qwen2.5-VL-3B is perfect on canvas/SVG relational object cases in the broad suite, but has lower hierarchy scores on tree/DOM cases and low table exactness. Figure~\ref{fig:constraint_profile} shows the first violated diagnostic constraint for each broad-suite prediction under the declared family order; the released score files retain the complete violated-constraint set for the same prediction. Qwen3-VL-4B has fewer residual violations, but the figure is not a general model-capacity claim: it is a profile of where each frozen checkpoint still violates the evaluator-defined binding constraints.

Task-specific survival gives a more local view. On graph cases, Qwen2.5-VL-3B satisfies identity/relation/role/exact constraints for 40/50 cases; four additional cases are invalid or unresolved. On tree and HTML-DOM hierarchy cases, the declared first violation is identity/group selection, after which surviving cases remain correct through hierarchy, role, and exact stages. On broad table cases, 23/50 cases satisfy entity selection and then remain exact. The ordered summary is useful for compact reporting, while the accompanying violated-constraint set prevents it from being read as a temporal model trace.

\subsection{RQ3: Failure Profiles Differ Across Frozen VLMs}

The Qwen3 result is not just a higher number in a leaderboard. The residual errors change character. The broad suite contains only three Qwen3 errors: one canvas relational case in which the model selected a same-attribute distractor instead of the object above the anchor, and two HTML-DOM hierarchy cases in which it chose the output node from the wrong group while preserving the output role. All three are valid candidate outputs. This profile differs from Qwen2.5-VL-3B, whose errors are concentrated in hierarchy/group selection, graph identity or invalid endpoints, and table entity/column binding. Qwen2.5-VL-3B is therefore a useful diagnostic operating point: validity is already high, but the residual population is large enough to expose structural patterns. Qwen3-VL-4B supplies a near-saturated contrast in which the same diagnostic vocabulary remains defined but yields few residual cases. ExBind makes this comparison possible because both models are scored through the same latent-stage vocabulary.

The supporting LLaVA-Phi-3-mini run provides a deliberately non-saturated cross-family profile under the same interface. It is valid on 220/250 broad cases and exact on 154/250; on the targeted table split it is valid on 67/120 and exact on 46/120. Among its table errors, 53 are invalid or unresolved and 21 are valid correct-row/wrong-column selections, classified as wrong-attribute errors under the table-specific diagnostic order. This matches the structural form of the Qwen2.5-VL-3B table profile, but with substantially lower validity and a different incidence of invalid outputs. We therefore use it as supporting evidence about cross-family validity and residual-profile coverage rather than as a matched conditional accuracy comparison. Capability-level opportunities and corresponding first-violation rates are reported with explicit applicability denominators in Appendix~\ref{app:reproducibility}; they are not pooled into a visual comparison because the dimensions apply to different representation subsets.

\subsection{RQ4: Table Binding Localizes a Consistent Column-Misbinding Pattern}

The strongest diagnostic case study is the targeted table split. Qwen2.5-VL-3B is valid on every table case but exact on only 63/120 (52.5\%). For every table error, the first violated diagnostic constraint under the declared row--column--cell order is \texttt{table\_column}: row resolution succeeds, but the selected cell is in the wrong attribute column. The full task-specific violation set also records the failed cell intersection. Qwen3-VL-4B is exact on all 120 table cases.

The structural conditions should not be read as a monotonic axis. They vary jointly in rows, candidates, constraints, source length, duplicate values, and gold-rank distribution. Qwen2.5-VL-3B exact accuracy is 0.275 on compact single-condition tables, 0.825 on conjunctive-condition tables, and 0.475 on high-candidate duplicate-value tables. A confound audit shows candidate-position effects: accuracy is 0.857 when the gold candidate is rank 1, 0.222 at rank 9, and 0.000 for sparse ranks 13 and above. Target column and template family are balanced within each condition, so the anomaly is not explained by all surface factors.

A fixed 20-case adjudication pack supports the stage interpretation. It includes all three table conditions, 14 Qwen2.5-VL-3B failures, and 6 correct controls. All 14 failures are valid candidate selections in the correct row but wrong column. The stage decomposition identifies the semantic form of the error as column misbinding. The confound audit shows that the frequency of that error is also sensitive to interface factors such as candidate position. This is a localized executable mismatch pattern, not evidence for a unique internal model mechanism.

Candidate order is a nuisance variable of the executable interface, not a latent binding capability: it changes the access path to an unchanged candidate set while leaving the observation, instruction, latent target, and gold reference fixed. We therefore interpret order sensitivity as interface sensitivity and report it as a measurement result. Stability of the first violated constraint under this nuisance supports the diagnostic error-type description, but does not identify the causal source of column binding.

A matched interface ablation adds only explicit per-candidate \texttt{column} metadata to the same 120 cases, preserving observation, instruction, candidate order, gold, and case IDs. Exact accuracy changes from 63/120 (0.525) to 71/120 (0.592), with 15 paired improvements and 7 regressions (Table~\ref{tab:tablediag}). This provides evidence that the observed column-binding failures are sensitive to candidate-interface specification; it is not evidence of a reliable accuracy improvement or a unique causal mechanism.

\begin{figure*}[t]
\centering
\includegraphics[width=0.96\linewidth]{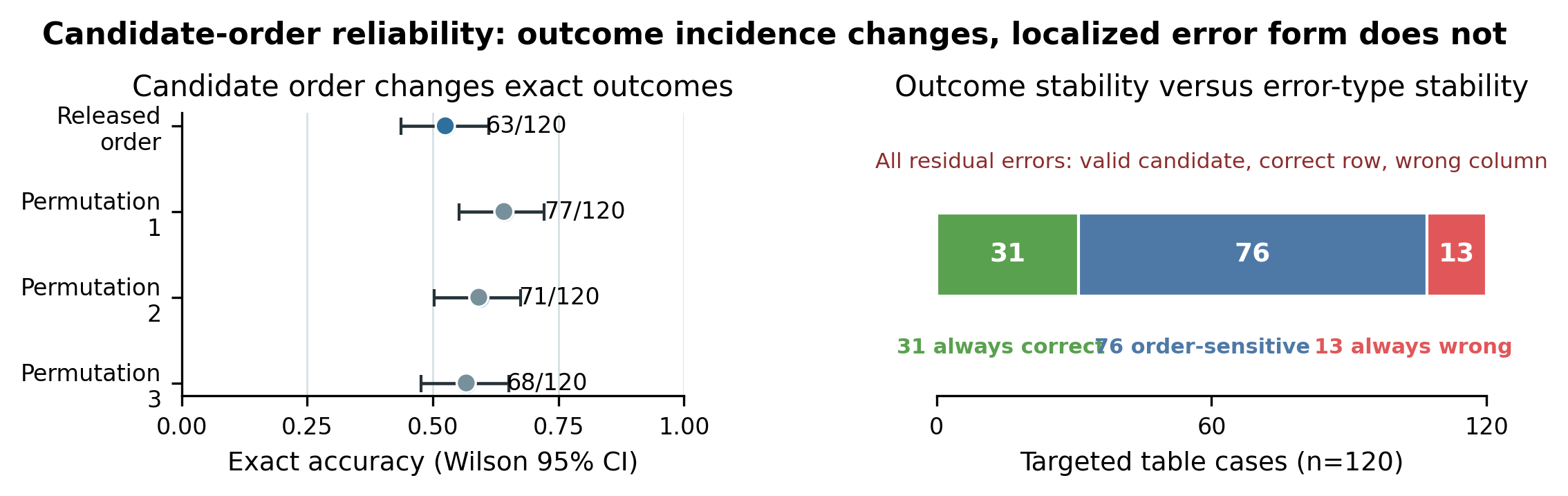}
\caption{Candidate-order reliability control on the same 120 targeted table cases. \textbf{Left:} exact accuracy varies across the released order and three deterministic candidate permutations; points show exact accuracy and error bars show Wilson 95\% intervals. \textbf{Right:} 76 cases change exact outcome across orders, while 31 remain correct and 13 remain wrong under all four interfaces. For each order, every residual error is a valid candidate selection from the correct row and wrong attribute column. Thus candidate order changes outcome incidence, whereas the evaluator-localized error form is stable.}
\label{fig:candidate_order_reliability}
\end{figure*}

We therefore ran one additional candidate-order control on the same 120 table cases. For each case we produced three deterministic permutations that preserve the latent record, observation, instruction, candidate set, and gold target; only candidate order changes. Qwen2.5-VL-3B exact accuracy is 77/120, 71/120, and 68/120 under the three permutations, compared with 63/120 under the original order. The paired mean delta over permutations is +7.5 percentage points with a 95\% paired bootstrap interval of [-1.1, 16.1]. The bootstrap unit is the 120 parent case IDs, with three repeated measurements per case; the 360 predictions are not treated as independent cases. This interval crosses zero, so the permutations are not interpreted as a reliable accuracy improvement. Figure~\ref{fig:candidate_order_reliability} makes the measurement result explicit: 76/120 cases are order-sensitive, whereas every residual error under every condition remains a valid correct-row/wrong-column selection. Candidate order is therefore a nuisance variable: it changes failure incidence and case identity while preserving the localized error form. Outcome stability is not diagnostic-type stability, and neither observation establishes a causal mechanism for column misbinding.

\begin{table*}[t]
\caption{Targeted table diagnostic and matched column-metadata ablation. Exact intervals are Wilson 95\% intervals; the ablation delta interval is paired bootstrap.}
\label{tab:tablediag}
\centering
\small
\begin{tabular*}{\linewidth}{@{\extracolsep{\fill}}p{0.24\linewidth}rrrrp{0.24\linewidth}@{}}
\toprule
Condition & Cases & Validity & Exact & Interval & Notes\\
\midrule
Original table interface & 120 & 1.000 & 63/120 & [0.436, 0.612] & all errors first violate column\\
Explicit column metadata & 120 & 1.000 & 71/120 & [0.502, 0.675] & 15 improve, 7 regress\\
Paired delta & 120 & -- & +8/120 & [-0.008, 0.142] & paired bootstrap\\
20-case adjudication & 20 & 1.000 & 6/20 & -- & 14 wrong-column errors\\
\bottomrule
\end{tabular*}
\end{table*}

\subsection{RQ5: Reliability and Cross-Representation Checks}

The hierarchy-depth diagnostic varies repeated-module depth from 2 to 4 while preserving deterministic hierarchy and role labels. Both evaluated models are exact on all 120 hierarchy diagnostic cases. This negative result is useful: depth alone is insufficient to challenge these models under this controlled construction, and ExBind does not declare every structural manipulation difficult.

Cross-format results are supporting evidence for the latent abstraction, not a universal representation-invariance claim. SVG--canvas relational pairs are fully stable for both models. Tree--DOM pairs show limited sensitivity: Qwen2.5-VL-3B has 0.80 semantic consistency over 25 paired groups, while Qwen3-VL-4B has 0.92. The paired transition analysis in Table~\ref{tab:cross_format_transitions} makes this result more specific. For Qwen2.5-VL-3B, five hierarchy mismatches persist across both tree and DOM surfaces, while five other groups change status across surfaces. Qwen3-VL-4B has two tree-correct to DOM-identity transitions. These paired outcomes show that some mismatches are latent-structure stable while others are surface-sensitive.

\begin{table}[H]
\caption{Cross-format semantic consistency on explicitly paired latent groups. Surface identifiers are not required to match.}
\label{tab:cross_format_results}
\centering
\scriptsize
\begin{tabular*}{\linewidth}{@{\extracolsep{\fill}}lrrr@{}}
\toprule
Pair & Groups & Qwen2.5 & Qwen3\\
\midrule
SVG--Canvas2D & 25 & 1.00 & 1.00\\
Tree--HTML-DOM & 25 & 0.80 & 0.92\\
\bottomrule
\end{tabular*}
\end{table}

The ordinary-contract LLaVA-1.5-7B-HF pilot is retained in Appendix~\ref{app:nonqwen} as a protocol-boundary result: its 7/120 targeted-table validity is too low for a useful conditional profile. LLaVA-Phi-3-mini improves broad and table coverage enough to expose a non-Qwen residual profile, but its 55.8\% table validity still prevents a primary matched comparison. The two Qwen checkpoints therefore remain the primary model comparison.

The legacy bridge tests whether the taxonomy applies beyond newly generated cases. A fixed 300-row eligible EVG/SEB subset maps established executable SVG binding outputs into the ExBind stage vocabulary without rerunning inference. It spans object, endpoint, group, and edge/rejection tasks and contains 152 no-failure rows and 148 identity-stage failures. Appendix~\ref{app:legacy} provides the corresponding benchmark-only context from the earlier full paper: ambiguity, perturbation, external-transfer, and task-conditioned failure analyses. This supports the narrow claim that the failure vocabulary can be applied to existing executable binding data; it is not a broad transfer result.

\section{Limitations}

ExBind isolates the referent-to-executable correspondence layer and therefore does not measure error accumulation over a complete multi-turn coding trajectory, patch generation, or execution. The primary model comparison uses two related Qwen-family checkpoints; LLaVA-Phi-3-mini provides supporting non-Qwen evidence, but its 55.8\% targeted-table validity limits matched conditional comparison, while the older LLaVA pilot remains a protocol-boundary result. The controlled suite is designed for identifiable structural diagnosis rather than population-scale ranking. Rendered SVG and canvas cases address the visual setting more directly, while serialized DOM, tree, graph, and table cases expose structural factors that would otherwise be difficult to attribute; the set is therefore not a uniform rendered-image evaluation. Cross-format pairing is limited to compatible compiler pairs and 50 latent groups, and the existing-task check is a fixed 20-error sample rather than an estimate of real-agent prevalence. Candidate serialization is part of the interface and can change case outcomes, so comparisons require the released order or an explicitly matched alternative. Finally, the first violated constraint is an ordered evaluator summary; full violated-constraint sets are released alongside it, and neither is a model reasoning trace or causal mechanism.

\section{Conclusion}

ExBind isolates visual/semantic-to-executable correspondence as a structured diagnostic object rather than a single exact-match bit. By separating latent targets, surface references, and deterministic structural constraints, it distinguishes invalid output, wrong identity, wrong relation, wrong hierarchy, wrong role, structured-selection errors, and final-reference errors. Frozen-model results show that high candidate validity can hide executable mismatches, while the table diagnostic exposes a consistent correct-row/wrong-column pattern under the declared decomposition. Candidate-order controls further show that interface serialization can alter case-level outcomes without necessarily changing the localized error form, making outcome stability and diagnostic-type stability distinct measurement properties. The paired latent construction extends this analysis across compatible surfaces, with both persistent and surface-sensitive transitions. Together, these results define a controlled instrument for measuring executable correspondence; full editing trajectories, model-internal mechanisms, and future intervention methods remain outside its scope.

\section*{Reproducibility Statement}

All paper results correspond to the ExBind v2.0 release manifest in the artifact directory. The benchmark schema and evaluation code are in the source tree, and the broad suite, targeted diagnostic, table audit, adjudication pack, explicit-column ablation, cached predictions, existing-task coverage annotations, and legacy bridge subset are stored as versioned artifacts. The repository-level \texttt{REPRODUCE\_EXBIND.md} and Appendix~\ref{app:reproducibility} list tested command lines for generation, frozen-model execution, deterministic scoring, aggregation, figure rendering, and paper compilation.

\onecolumn
\appendix

\section{Reproducibility and Benchmark Specification}
\label{app:reproducibility}

\subsection{Data Generation}

The broad diagnostic suite is generated by \texttt{scripts/run\_exbind\_v2\_cross\_format.py} with seed \texttt{20260823}, generator version \texttt{cross\_format.interface\_gate.v1}, \texttt{n\_per\_family=50}, and \texttt{n\_cross\_format\_groups=50}. The targeted structural diagnostic is generated by \texttt{scripts/generate\_exbind\_v2\_hard\_pilot.py} with seed \texttt{20260824}, generator version \texttt{exbind\_v2.hard\_pilot.v1}, and 40 examples per family-level cell. Relational latent instances are compiled into canvas and SVG surfaces; hierarchy instances into tree and HTML-DOM surfaces; graph instances into directed endpoint candidates; and table instances into HTML-table cell candidates. Candidate inventories are constructed from the compiler output. Gold references are derived through the latent-to-surface mapping, not by string matching against the instruction.

\paragraph{Algorithm 1: ExBind Case Generation.}
\begin{enumerate}
    \item Input representation family $f$ and seed $s$.
    \item Sample latent binding instance $G=(V,E,A,H,R,Y)$.
    \item Sample the target specification $Y$, including the canonical semantic target.
    \item Compile the observation $O_f=K_f(G)$.
    \item Construct executable candidate inventory $C_f$ from the surface structure.
    \item Derive surface gold $Y_f$ through the latent-to-surface mapping $M_f$.
    \item Derive deterministic stage labels from $G$, $C_f$, $Y$, $Y_f$, and $M_f$.
    \item Run integrity gates: unique gold, non-empty candidates, structural oracle, target-ID leakage, lexical-only baseline, type-only baseline, and candidate-order shortcut check.
    \item Reject the case if any required gate fails.
    \item Store the latent record, rendered surface case, candidate inventory, mappings, gold reference, and stage labels.
\end{enumerate}

\subsection{Benchmark Record Schema}

All generated and converted cases are represented as \texttt{BindingRecord} objects in \texttt{src/exbind/schema.py}. The fields are intentionally sparse: dimensions that do not apply to a task are null rather than scored as failures. A simplified table record is shown below.

\small
\begin{verbatim}
{
  "case_id": "hard_table_easy_0040__table",
  "latent_id": "hard_table_easy_0040",
  "format": "table",
  "instruction": "Select the status cell for the project
                  where owner Team B.",
  "observation": {"source_type": "html_table", "source": "<table>..."},
  "candidates": [
    {"candidate_id": "cell_r01_c00", "candidate_type": "cell",
     "semantic_id": "...row_1.owner"},
    {"candidate_id": "cell_r01_c01", "candidate_type": "cell",
     "semantic_id": "...row_1.status"}
  ],
  "gold": {"cell_id": "cell_r01_c01"},
  "binding": {
    "entity": "hard_table_easy_0040_row_1",
    "attribute": "status",
    "target_type": "cell",
    "executable_type": "cell_id"
  },
  "surface_mapping": {
    "hard_table_easy_0040_row_1.status": "cell_r01_c01"
  },
  "applicable_stages": [
    "table_row", "table_column", "table_cell", "exact"
  ]
}
\end{verbatim}
\normalsize

\subsection{Prediction and Diagnostic Scoring Contract}

The model is asked to return strict JSON only. It is never asked to produce chain-of-thought. A table prediction has the form:

\small
\begin{verbatim}
{"cell_id": "cell_r01_c00"}
\end{verbatim}
\normalsize

The evaluator parses the JSON, normalizes field aliases, checks candidate validity against the candidate inventory, maps the surface reference back to the canonical latent target, and emits a deterministic diagnostic score vector. For the table example above, selecting the owner cell in the correct row yields:

\small
\begin{verbatim}
{
  "table_row": 1,
  "table_column": 0,
  "table_cell": 0,
  "candidate_valid": 1,
  "executable_exact": 0
}
\end{verbatim}
\normalsize

Malformed JSON is parsed with the repository JSON extractor when possible; otherwise the prediction is treated as unresolved. Unknown candidate identifiers are invalid. For graph cases, both \texttt{source\_id} and \texttt{target\_id} must be endpoint candidate IDs, and reversing them is scored as a role/order failure rather than as an unordered object match.

\subsection{Prompt and Inference Specification}

The frozen-model pilot uses the same ordinary prompt builder for all listed checkpoints:

\small
\begin{verbatim}
You are evaluating a binding benchmark. Select only from
the candidate IDs.
Format: {format}
Instruction: {instruction}

Observation ({source_type}):
{serialized_source}

Candidates:
{candidate_json}

Return strict JSON only with this schema: {schema}
\end{verbatim}
\normalsize

The schema is \texttt{\{"target\_id": "..."\}} for ordinary single-target cases, \texttt{\{"source\_id": "...", "target\_id": "..."\}} for graph endpoint-pair cases, \texttt{\{"dom\_id": "..."\}} for HTML-DOM cases, and \texttt{\{"cell\_id": "..."\}} for table cases. Candidate rows include \texttt{candidate\_id}, \texttt{candidate\_type}, and bounding boxes when available. Graph rows additionally include endpoint role metadata. The explicit-column ablation adds only a \texttt{column} field to table candidate rows; observation, instruction, candidate order, gold, and case IDs are unchanged. The candidate-order control changes only the order of the same candidate rows using fixed seeds \texttt{2026082401}, \texttt{2026082402}, and \texttt{2026082403}.

Table~\ref{tab:model_protocol} specifies the frozen-model interface used for the paper results. The same ordinary contract is applied to both primary checkpoints and the supporting LLaVA-Phi pilot, so comparisons are made under a fixed reference-output protocol rather than task-specific prompt tuning. When a rendered image is present, the loader passes it through the model processor chat template; otherwise the observation is provided as serialized SVG, HTML, DOM, graph, or table source text. Historical cached predictions store the model path and decoding configuration. The paper-polish release adds \texttt{frozen\_inference\_manifest.json}, which records recovered checkpoint, package, device, decoding, prompt, parser, and unrecoverable historical-run fields.

\begin{table}[H]
\caption{Frozen-model inference protocol.}
\label{tab:model_protocol}
\centering
\small
\begin{tabular*}{\linewidth}{@{\extracolsep{\fill}}p{0.22\linewidth}p{0.74\linewidth}@{}}
\toprule
Field & Protocol\\
\midrule
Checkpoints & Qwen2.5-VL-3B-Instruct, Qwen3-VL-4B-Instruct, and LLaVA-Phi-3-mini\\
Decoding & Greedy decoding with sampling disabled\\
Generation budget & Maximum 96 new tokens\\
Random seed & 42 for the evaluation wrapper\\
Prompt & Shared strict-reference template in Appendix~\ref{app:reproducibility}\\
Candidate serialization & JSON candidate rows derived from the compiler output; graph rows include endpoint role metadata\\
Output contract & Strict JSON executable reference matching the task family\\
Parsing & JSON extraction followed by field-alias normalization and candidate-inventory validation\\
Invalid outputs & Malformed or unresolved outputs are invalid; unknown identifiers are invalid candidate selections\\
\bottomrule
\end{tabular*}
\end{table}

The supporting LLaVA-Phi-3-mini pilot uses the same prompt builder, candidate serialization, and unconstrained greedy decoding as the frozen Qwen interface. Its broad and targeted predictions, diagnostic scores, and run manifests are included in the versioned release artifact bundle; task-specific precedence-rescored copies used for the reported failure taxonomy are retained alongside them. The older ordinary-contract LLaVA-1.5-7B-HF boundary pilot is retained as an appendix artifact. The separate finite candidate-constrained adapter is not comparable and is excluded from the paper's model results.

\subsection{Candidate-Order Control Artifacts}

The candidate-order control is included in the versioned release artifact bundle. It contains the generated permutation records, the order mapping, raw predictions, deterministic diagnostic scores, per-condition summaries, per-case transitions, transition matrix, gold-rank summaries, and the experiment manifest. The mapping file records, for every case and permutation, the original candidate order, permuted candidate order, original gold rank, and permuted gold rank. The control uses the same Qwen2.5-VL-3B checkpoint and decoding protocol as the frozen targeted table diagnostic.

\begin{table}[H]
\caption{Candidate-order control on the same 120 targeted table cases. Only candidate ordering changes. Exact intervals are Wilson 95\% intervals; the paired delta interval is computed by paired bootstrap over case IDs.}
\label{tab:candidate_order}
\centering
\small
\begin{tabular*}{\linewidth}{@{\extracolsep{\fill}}lrrrrp{0.28\linewidth}@{}}
\toprule
Condition & Cases & Validity & Exact & Exact 95\% CI & Error profile\\
\midrule
Original order & 120 & 1.000 & 63/120 & [0.436, 0.612] & 57/57 errors wrong column\\
Permutation 1 & 120 & 1.000 & 77/120 & [0.553, 0.722] & 43/43 errors wrong column\\
Permutation 2 & 120 & 1.000 & 71/120 & [0.502, 0.675] & 49/49 errors wrong column\\
Permutation 3 & 120 & 1.000 & 68/120 & [0.477, 0.652] & 52/52 errors wrong column\\
\midrule
Paired case IDs & 120 & -- & +7.5 pp & [-1.1, 16.1] & three permutations per case; 76 order-sensitive cases\\
\bottomrule
\end{tabular*}
\end{table}

\subsection{Uncertainty Estimates}

For the broad diagnostic suite, Qwen2.5-VL-3B exact accuracy is 191/250, or 76.4\% with a Wilson 95\% interval of [70.8, 81.2]. Qwen3-VL-4B exact accuracy is 247/250, or 98.8\% [96.5, 99.6]. On the targeted table diagnostic, Qwen2.5-VL-3B is 63/120, or 52.5\% [43.6, 61.2], while Qwen3-VL-4B is 120/120, or 100.0\% [96.9, 100.0]. The explicit-column metadata ablation changes Qwen2.5-VL-3B from 63/120 to 71/120. The paired bootstrap estimate is +6.7 percentage points with a 95\% interval of [-0.8, 14.2], so we describe it as interface sensitivity rather than as a reliable improvement claim. The candidate-order control has a paired mean delta of +7.5 percentage points over three permutations with a 95\% paired bootstrap interval of [-1.1, 16.1].

\subsection{Constraint Sets and Declared-Order Summaries}

For every prediction, the release stores both the stage-level score vector and the full set of applicable structural constraints that are violated. Exact-reference mismatch is retained as a separate terminal score because it is necessarily zero for every non-exact prediction; including it in every set would obscure the structural differences among errors. The first-violation label is computed only after fixing a task-family order supplied by the benchmark definition. For example, the targeted table order is row/entity $\rightarrow$ column $\rightarrow$ cell $\rightarrow$ exact reference, while a tree/DOM order uses identity $\rightarrow$ hierarchy $\rightarrow$ role $\rightarrow$ exact reference.

This distinction matters because several structural scores can co-vary after one surface selection is wrong. We therefore do not claim that first-violation labels are invariant to all alternative decompositions or that they reveal temporal model operations. The deterministic robustness analysis enumerates admissible permutations of internal structural stages while keeping candidate validity first and executable exactness last; it writes both the declared first violation and the set of possible first constraints to the versioned release artifact bundle. The main paper uses the declared order only as a concise, reproducible summary, and retains the full sets for any analysis that should not depend on ordering.

\subsection{Paired Cross-Format Diagnostic Transitions}

The paired transition analysis consumes the frozen score vectors and explicit latent groups; it requires no new model inference. For each pair, it maps both surface predictions to their declared first-violation label and stores the transition together with the two surface identifiers. The SVG--canvas pairs are all none-to-none for both primary checkpoints. For tree--DOM groups, the primary transition counts are reported in Table~\ref{tab:cross_format_transitions}. The paired rows and matrix are stored beside the constraint-set analysis. They support the limited conclusion that the common diagnostic vocabulary can describe both persistent and surface-sensitive mismatches across compatible compilers.

\begin{table}[H]
\caption{Tree--DOM paired diagnostic transitions (25 groups). C denotes no violated constraint and I denotes declared first-violation identity. Counts are not an aggregate leaderboard score.}
\label{tab:cross_format_transitions}
\centering
\small
\begin{tabular*}{\linewidth}{@{\extracolsep{\fill}}lrrrr@{}}
\toprule
Model & C $\rightarrow$ C & C $\rightarrow$ I & I $\rightarrow$ C & I $\rightarrow$ I\\
\midrule
Qwen2.5-VL-3B & 15 & 2 & 3 & 5\\
Qwen3-VL-4B & 23 & 2 & 0 & 0\\
\bottomrule
\end{tabular*}
\end{table}

\subsection{Construct Applicability and Opportunity-Normalized Profiles}

The applicability audit makes the benchmark's denominators explicit. Table~\ref{tab:construct_coverage} reports the number of broad-suite cases for which each evaluator-defined constraint is scored; a non-applicable dimension is excluded rather than treated as an error. Table~\ref{tab:normalized_profile} reports the nonzero first-violation counts together with these opportunity denominators. These rates describe diagnostic localization under the fixed decomposition, not causal probabilities or model-internal reasoning stages.

\begin{table}[H]
\caption{Applicability opportunities in the 250-case broad suite.}
\label{tab:construct_coverage}
\centering
\scriptsize
\begin{tabular*}{\linewidth}{@{\extracolsep{\fill}}lrrl@{}}
\toprule
Constraint & Applicable & Fraction & Representations\\
\midrule
Candidate validity & 250 & 1.000 & all six\\
Identity & 250 & 1.000 & all six\\
Attribute & 200 & 0.800 & canvas, DOM, SVG, table, tree\\
Anchor & 75 & 0.300 & canvas, SVG\\
Relation & 125 & 0.500 & canvas, graph, SVG\\
Hierarchy & 75 & 0.300 & DOM, tree\\
Ordered role & 125 & 0.500 & graph, DOM, tree\\
Structured selection & 50 & 0.200 & table\\
Exact reference & 250 & 1.000 & all six\\
\bottomrule
\end{tabular*}
\end{table}

\begin{table}[H]
\caption{Opportunity-normalized first-violation profile. Rates use the applicable denominator for each evaluator-defined constraint; ``none'' denotes no violated constraint.}
\label{tab:normalized_profile}
\centering
\scriptsize
\begin{tabular*}{\linewidth}{@{\extracolsep{\fill}}lrrr@{}}
\toprule
Constraint & Opportunities & Qwen2.5 & Qwen3\\
\midrule
Candidate validity & 250 & 4 (0.016) & 0 (0.000)\\
Identity & 250 & 28 (0.112) & 3 (0.012)\\
Table column & 50 & 27 (0.540) & 0 (0.000)\\
No violated constraint & 250 & 191 (0.764) & 247 (0.988)\\
\bottomrule
\end{tabular*}
\end{table}

\subsection{Provenance and Reproduction Interface}

The intended reproduction chain is:
\begin{center}
\small
data generation $\rightarrow$ frozen prediction $\rightarrow$ scoring $\rightarrow$ paper tables.
\end{center}
The release exposes this chain as four deterministic stages. Table~\ref{tab:repro_pipeline} gives the tested entry points, inputs, and outputs. Appendix~\ref{app:reproduction_commands} and the repository-level reproduction guide provide the exact command lines; each stage consumes a versioned artifact and writes counts, configuration, and hashes without changing the frozen inputs.

\begin{table}[H]
\caption{Reproduction interface for paper-facing results.}
\label{tab:repro_pipeline}
\centering
\scriptsize
\begin{tabular*}{\linewidth}{@{\extracolsep{\fill}}p{0.13\linewidth}p{0.19\linewidth}p{0.23\linewidth}p{0.24\linewidth}@{}}
\toprule
Stage & Entry point & Input & Output\\
\midrule
Generate broad suite & Cross-format generator & Seed 20260823; interface-gate generator config & Binding records, cross-format groups, benchmark manifest\\
Generate targeted suite & Hard-pilot generator & Seed 20260824; table and hierarchy diagnostic config & Binding records, gate summary, hard-pilot manifest\\
Frozen prediction & Frozen-model evaluator & Binding records, optional cross-format groups, checkpoint identifier, decoding protocol & Prediction JSONL, diagnostic-score JSONL, run manifest\\
Paper aggregation & Summary/table builder & Diagnostic-score files, cross-format groups, legacy bridge subset & Accuracy tables, stage profiles, consistency summaries, figure data\\
\bottomrule
\end{tabular*}
\end{table}

The ExBind v2.0 manifest records the artifact set used for the paper. Representative hash prefixes are listed in Table~\ref{tab:release_hashes}; the release manifest contains the complete SHA256 values for the benchmark, prediction, ablation, and bridge artifacts.

\begin{table}[H]
\caption{Representative release artifacts.}
\label{tab:release_hashes}
\centering
\small
\begin{tabular*}{\linewidth}{@{\extracolsep{\fill}}p{0.42\linewidth}p{0.52\linewidth}@{}}
\toprule
Artifact & SHA256 prefix\\
\midrule
Broad-suite binding records & \texttt{e9925d0ef231e404}\\
Targeted-suite binding records & \texttt{6d779aa8e898c50c}\\
Existing-task coverage manifest & \texttt{5eb6db8d63b697d4}\\
\bottomrule
\end{tabular*}
\end{table}

\subsection{Tested Reproduction Commands}
\label{app:reproduction_commands}

The following commands are the paper-facing entry points tested for the release-readiness pass. They are shown with repository-relative paths and write to fresh output directories. The generator smoke tests produced five broad records and six targeted records when run with one example per family/level; the full paper configuration uses the seeds and counts specified in Appendix~\ref{app:reproducibility}.

{\scriptsize
\begin{verbatim}
export PYTHONPATH=src:.
python3 scripts/run_exbind_v2_cross_format.py \\
  --out artifacts/reproduction/exbind_v2_cross_format \\
  --n-per-family 50 --n-cross-format-groups 50 \\
  --seed 20260823 --surface-variant interface_gate_v1

python3 scripts/generate_exbind_v2_hard_pilot.py \\
  --out artifacts/reproduction/exbind_v2_hard_pilot \\
  --per-level 40 --seed 20260824

python3 scripts/summarize_exbind_v2_frozen_model_pilot.py \\
  --groups artifacts/exbind_v2/cross_format_interface_gate_v1/cross_format_groups.jsonl \\
  --model Qwen2.5-VL-3B=artifacts/exbind_v2/frozen_model_pilot_interface_gate_v1/qwen25_3b \\
  --model Qwen3-VL-4B=artifacts/exbind_v2/frozen_model_pilot_interface_gate_v1/qwen3_4b \\
  --out artifacts/reproduction/frozen_model_summary

python3 scripts/run_exbind_v2_evidence_completion.py \\
  --out artifacts/reproduction/evidence_completion

python3 scripts/render_exbind_v2_arr_figures.py \\
  --paper-dir artifacts/reproduction/paper_assets \\
  --failure-csv "$FAILURE_CSV"

python3 scripts/paper_figures/build_exbind_v2_candidate_order_reliability.py \\
  --artifact-dir artifacts/paper_assets/exbind_v2_arr_polish/candidate_order_robustness \\
  --out-dir Exbind/arr2026_october_benchmark/figures

cd Exbind/arr2026_october_benchmark
latexmk -pdf -interaction=nonstopmode -halt-on-error \\
  -outdir=build exbind_benchmark.tex
\end{verbatim}
}

Frozen-model inference uses the same evaluator entry point after replacing the model argument with a local checkpoint path, for example:
{\scriptsize
\begin{verbatim}
python3 scripts/run_exbind_v2_frozen_model_pilot.py \\
  --records artifacts/reproduction/exbind_v2_cross_format/binding_records.jsonl \\
  --groups artifacts/reproduction/exbind_v2_cross_format/cross_format_groups.jsonl \\
  --out artifacts/reproduction/qwen25_broad \\
  --model "$QWEN25_VL_3B_PATH" --model-name Qwen2.5-VL-3B \\
  --model-loader qwen --seed 42 --max-new-tokens 96 --write-prompts
\end{verbatim}
}
The release does not distribute model weights. The complete inference fields, including unrecoverable historical fields, are listed in the frozen inference manifest. The tested command log and smoke manifest are stored under \path{artifacts/paper_assets/exbind_v2_release_readiness}.

\subsection{Benchmark Card}

\begin{table}[H]
\caption{ExBind-v2 benchmark card.}
\label{tab:benchmark_card}
\centering
\small
\begin{tabular*}{\linewidth}{@{\extracolsep{\fill}}p{0.20\linewidth}p{0.68\linewidth}@{}}
\toprule
Field & Specification\\
\midrule
Intended use & Diagnostic evaluation of executable reference binding and failure localization\\
Unit of evaluation & One instruction, observation, candidate inventory, latent target, and executable gold\\
Representations & SVG, DOM, canvas, tree, graph, table\\
Binding dimensions & Identity, attribute, relation, hierarchy, role/order, structured selection, executable exactness\\
Model output & Strict JSON executable reference; no chain-of-thought required\\
Metrics & Candidate validity, exact accuracy, dimension accuracy, violated-constraint set, survival, declared-order first violation, cross-format consistency\\
Primary splits & 250 broad diagnostic cases and 240 targeted structural cases\\
Data source & Deterministically generated latent records compiled into six surface representation families; inherited EVG/DOM data are separate legacy resources\\
Model baselines & Primary frozen Qwen2.5-VL-3B and Qwen3-VL-4B; supporting LLaVA-Phi-3-mini and appendix-only LLaVA-1.5-7B-HF pilot\\
Validation & Unique gold, structural oracle, target-ID leakage, lexical/type baselines, candidate-order diagnostics\\
Reference controls & Structural oracle for gold recoverability and random-candidate expectations for chance calibration\\
Release artifact & Original manifest preserved at \path{artifacts/exbind_v2/exbind_v2_0_release_manifest.json}; reconciled copy and evidence hashes at \path{artifacts/paper_assets/exbind_v2_evidence_completion}\\
License status & Core code is MIT-scoped and newly generated core records are CC BY 4.0-scoped; inherited raw assets with unresolved upstream terms are excluded from the core release\\
Known limitations & Two primary Qwen checkpoints; supporting LLaVA-Phi-3-mini has low table validity; controlled/synthetic cases; candidate serialization sensitivity\\
Version & \texttt{ExBind-v2.0}\\
\bottomrule
\end{tabular*}
\end{table}

\subsection{Protocol-Boundary Non-Qwen Pilot}
\label{app:nonqwen}

We retain two ordinary-contract non-Qwen pilots to document model-family variation beyond the primary Qwen checkpoints. LLaVA-Phi-3-mini uses the same prompt builder, candidate serialization, and unconstrained greedy decoding; it obtains 220/250 valid and 154/250 exact outputs on the broad suite, and 67/120 valid and 46/120 exact outputs on the targeted table suite. Its table errors include 21 valid correct-row/wrong-column selections, classified as wrong-attribute errors under the table-specific diagnostic order, alongside 53 invalid or unresolved outputs. The older LLaVA-1.5-7B-HF pilot obtains 250/250 valid and 75/250 exact broad outputs, but only 7/120 valid and 4/120 exact targeted-table outputs; it remains a protocol-boundary artifact. Neither pilot is included in the primary matched model-profile comparison because the table validity condition differs substantially from the two Qwen checkpoints. A separate finite candidate-constrained adapter is excluded entirely because it changes the output interface.

\begin{table}[H]
\caption{Frozen ordinary-contract non-Qwen pilot, reported only as a protocol boundary.}
\label{tab:llava_support}
\centering
\small
\begin{tabular*}{\linewidth}{@{\extracolsep{\fill}}lrrr@{}}
\toprule
Model & Broad valid/exact & Table valid/exact & Role in paper\\
\midrule
LLaVA-Phi-3-mini & 220/250 ; 154/250 & 67/120 ; 46/120 & supporting cross-family profile\\
LLaVA-1.5-7B-HF & 250/250 ; 75/250 & 7/120 ; 4/120 & appendix-only boundary\\
\bottomrule
\end{tabular*}
\end{table}

\subsection{Legacy Diagnostic Resources}
\label{app:legacy}

The inherited ExBind-Bench inventory is retained as historical benchmark material and as a source for cached-output bridge analysis. Table~\ref{tab:legacy_inventory} lists the full 5.9K Stage-A inventory. Supporting non-Qwen runs and the separate non-comparable adapter artifacts are stored separately from the primary Qwen prediction manifests and are not used to redefine any primary split or aggregate score.

\begin{table*}[t]
\caption{Inherited ExBind-Bench inventory. Counts are Stage-A executable-reference binding instances.}
\label{tab:legacy_inventory}
\centering
\scriptsize
\begin{tabular*}{\linewidth}{@{\extracolsep{\fill}}p{0.13\linewidth}p{0.12\linewidth}p{0.06\linewidth}p{0.07\linewidth}p{0.20\linewidth}p{0.27\linewidth}@{}}
\toprule
Split & Domain & Cases & Lineages & Task mix & Benchmark role\\
\midrule
EVG-Bench & controlled SVG & 800 & 800 & edge, endpoint, group, object & clean in-domain executable binding\\
EVG-Bench-Perturbed & perturbed SVG & 3,200 & 800 & edge, endpoint, group, object & lineage-preserving shortcut and robustness tests\\
Novel-SVG & held-out SVG & 300 & 300 & endpoint, object & held-out template transfer\\
SEB-hard & external SVG & 300 & 300 & endpoint, object & SVGEditBench-derived hard transfer\\
SEB-random & external SVG & 300 & 300 & object & SVGEditBench-derived random transfer\\
SEB-perturbed & external SVG & 500 & 100 & endpoint, object & external perturbation stress tests\\
MiniWoB-Binding & HTML/DOM & 500 & 500 & DOM-reference selection & transfer from SVG identifiers to DOM elements\\
\midrule
Total & SVG/DOM & 5,900 & -- & -- & inherited Stage-A benchmark inventory\\
\bottomrule
\end{tabular*}
\end{table*}

The earlier multi-panel legacy characterization is retained as a release artifact rather than included in the submission PDF. Its role is provenance for inherited assets, not a second primary evaluation narrative.

\subsection{Existing-Task Taxonomy Coverage}
\label{app:existing_coverage}

To test whether the controlled vocabulary can also describe existing executable-task errors, we fix a 20-error sample from cached EVG, SVGEditBench-derived, and MiniWoB DOM runs. A single manual adjudication compares each stored instruction, candidate/gold record, and cached prediction against the released taxonomy. The protocol and item-level annotations are versioned separately from all historical inputs. Table~\ref{tab:existing_coverage} reports coverage rather than prevalence: all sampled errors are covered, five multi-edit instructions are ambiguous at the task-unit level, and no example requires a new category.

\begin{table}[H]
\caption{Fixed existing-task error sample used for taxonomy-coverage checking. ``Ambiguous'' denotes task-unit ambiguity in a multi-edit instruction, not an ambiguous executable gold for the stored error.}
\label{tab:existing_coverage}
\centering
\small
\begin{tabular*}{\linewidth}{@{\extracolsep{\fill}}lrrr@{}}
\toprule
Source & Errors & Taxonomy covered & Ambiguous\\
\midrule
EVG & 8 & 8 & 0\\
SVGEditBench-derived & 10 & 10 & 5\\
MiniWoB DOM & 2 & 2 & 0\\
\midrule
Total & 20 & 20 & 5\\
\bottomrule
\end{tabular*}
\end{table}

The observed categories are 16 wrong-identity errors, two wrong-role/order errors, one wrong-relation error, and one wrong-type error. This supports only the narrow claim that the controlled taxonomy extends to this fixed sampled subset of existing executable tasks. It neither estimates real-agent failure frequencies nor validates the taxonomy for all open-ended editing trajectories. The coverage artifact is included in the versioned release artifact bundle.

\end{document}